\documentclass{article}
\usepackage{times}
\usepackage{graphicx} 
\usepackage{subfigure} 
\usepackage{footnote}
\usepackage{amsfonts}

\usepackage{natbib}
\usepackage{comment}
\usepackage{multirow}

\usepackage{algorithm}
\usepackage{algorithmic}
\usepackage{amsmath}

\usepackage[compact]{titlesec}
\titlespacing{\section}{0pt}{0.5ex}{0.3ex}
\titlespacing{\subsection}{0pt}{0.2ex}{0ex}
\titlespacing{\subsubsection}{0pt}{0.1ex}{0ex}

\usepackage{float}

\floatstyle{plain} 
\newfloat{code}{H}{myc}

\usepackage{iclr2015,times}
\iclrconference

\usepackage{paralist}

\makeatletter
\ifcase \@ptsize \relax
  \newcommand{\miniscule}{\@setfontsize\miniscule{4}{5}}
\or
  \newcommand{\miniscule}{\@setfontsize\miniscule{5}{6}}
\or
  \newcommand{\miniscule}{\@setfontsize\miniscule{5}{6}}
\fi
\makeatother

\newcommand {\aplt} {\ {\raise-.5ex\hbox{$\buildrel<\over\sim$}}\ }

\usepackage[symbol*]{footmisc}

\DefineFNsymbolsTM{myfnsymbols}{
  \textasteriskcentered *
  \textdagger    \dagger
  \textdaggerdbl \ddagger
  \textsection   \mathsection
  \textbardbl    \|%
  \textparagraph \mathparagraph
}%

\usepackage{hyperref}

\title{Recurrent Neural Network Regularization}

\begin{document} 


\author{
Wojciech Zaremba\footnote{}\footnotetext{Work done while the author was in Google Brain.} \\
New York University\\
\texttt{woj.zaremba@gmail.com} \\
\And
Ilya Sutskever, Oriol Vinyals \\
Google Brain \\
\texttt{\{ilyasu,vinyals\}@google.com} \\
}

\maketitle

\begin{abstract} 
  We present a simple regularization technique for Recurrent Neural
  Networks (RNNs) with Long Short-Term Memory (LSTM) units.  Dropout,
  the most successful technique for regularizing neural networks, does
  not work well with RNNs and LSTMs.  In this paper, we show how to
  correctly apply dropout to LSTMs, and show that it
  substantially reduces overfitting on a variety of tasks. These tasks
  include language modeling, speech recognition, image caption generation, and machine
  translation.
\end{abstract} 

{\textcolor{white} {\fontsize{0.001cm}{0.01em}\selectfont \footnotemark\footnotetext{Work done while the author was in Google Brain.} }}

\section{Introduction}

The Recurrent Neural Network (RNN) is neural sequence model that achieves state of the art
performance on important tasks that include language modeling
\cite{mikolov2012statistical}, speech recognition
\cite{graves2013speech}, and machine translation
\cite{kal13}.  It is known that successful applications of
neural networks require good regularization. Unfortunately, dropout
\cite{srivastava2013improving}, the most powerful regularization method
for feedforward neural networks, does not work well with
RNNs. As a result, practical applications of RNNs often
use models that are too small because large RNNs tend to overfit.  
Existing regularization methods give relatively small improvements for
RNNs \cite{graves2013generating}.
In this work, we show that dropout, when correctly used,
greatly reduces overfitting in LSTMs, and evaluate it on three different problems.

The code for this work can be found in \url{https://github.com/wojzaremba/lstm}.

\section{Related work}

Dropout \cite{srivastava2013improving} is a recently introduced
regularization method that has been very successful with
feed-forward neural networks.   While much work has extended dropout in  
various ways \cite{wang2013fast,wan2013regularization}, there has
been relatively little research in applying it to RNNs. The only
paper on this topic is by \citet{bayer2013fast}, who focuses on
``marginalized dropout'' \cite{wang2013fast}, a
noiseless deterministic approximation to standard dropout.
\citet{bayer2013fast} claim that conventional dropout does not work well
with RNNs because the recurrence amplifies noise, which in turn hurts learning. 
In this work, we show that this problem can be fixed by
applying dropout to a certain subset of the RNNs' connections.  As a result, RNNs
can now also benefit from dropout.


Independently of our work, \cite{pham2013dropout} developed the very same RNN 
regularization method and applied it to handwriting recognition.  We rediscovered
this method and demonstrated strong empirical results over a wide range of problems. 
Other work that applied dropout to LSTMs is \cite{pachitariu2013regularization}.

There have been a number of architectural variants of the RNN that
perform better on problems with long term dependencies
\cite{hochreiter1997long, graves2009novel, cho2014learning,
  jaeger2007optimization, koutnik2014clockwork, sundermeyer2012lstm}.  In this work, we
show how to correctly apply dropout to LSTMs, the most
commonly-used RNN variant; this way of applying dropout is likely to
work well with other RNN architectures as well. 

In this paper, we consider the following tasks: language modeling,
speech recognition, and machine translation.  Language modeling is the
first task where RNNs have achieved substantial success
\cite{mikolov2010recurrent, mikolov2011strategies,
  pascanu2013construct}.  RNNs have also been successfully used for
speech recognition \cite{robinson1996use, graves2013speech} and have
recently been applied to machine translation, where they are 
used for language modeling, re-ranking, or phrase modeling
\cite{devlin14,kal13,cho2014learning,chow1987byblos,mikolov2013exploiting}.

\section{Regularizing RNNs with LSTM cells}

In this section we describe the deep LSTM (Section \ref{sec:lstm}). Next, 
we show how to regularize them (Section \ref{sec:reg}), and explain
why our regularization scheme works.

We let subscripts denote timesteps and superscripts denote 
layers.  All our states are $n$-dimensional.  Let $h^l_t
\in \mathbb{R}^{n}$ be a hidden state in layer $l$ in timestep
$t$. Moreover, let $T_{n,m}:\mathbb{R}^{n} \rightarrow \mathbb{R}^{m}$
be an affine transform ($Wx + b$ for some $W$ and $b$).
Let $\odot$ be element-wise multiplication and let $h^0_t$ be an
input word vector at timestep $k$.  We use the activations $h^{L}_t$ to predict $y_t$,
since $L$ is the number of layers in our deep LSTM.

\subsection{Long-short term memory units}
\label{sec:lstm}

The RNN dynamics can be described using deterministic transitions
from previous to current hidden states. 
The deterministic state transition is a function
\begin{align*}
  &\text{RNN} : h^{l-1}_t, h^l_{t-1} \rightarrow h^l_t
\end{align*}

For classical RNNs, this function is given by
\begin{align*}
  h^l_t = f(T_{n,n}h^{l-1}_t + T_{n,n}h^l_{t-1}) \text{, where $f \in \{\mathrm{sigm}, \tanh\}$ }
\end{align*}

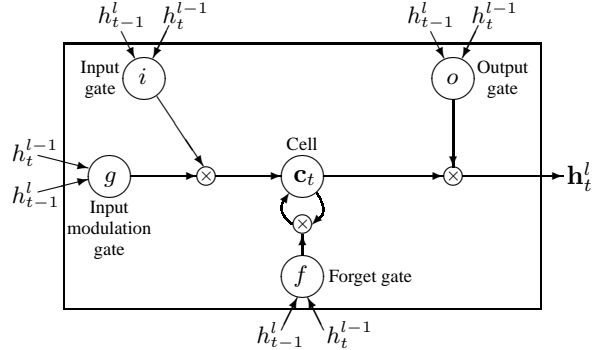
\begin{figure}
  \begin{center}
    \begin{picture}(200, 130)
      \put(0, 0){\framebox(180, 100){}}
      \put(90, 50){\circle{16}}
      \put(86.5, 48){$\mathbf c_t$}
      \put(84, 60){{\scriptsize Cell}}

      \put(90, 32){\circle{6.5}}
      \put(87.25, 30.5){{\tiny $\times$}}

      \put(90, 12){\circle{16}}
      \put(86.5, 10){{\small $f$}}
      \put(100, 10){{\scriptsize Forget gate}}

      \put(90, 20){\vector(0, 0){8}}

      \put(85, 44){\vector(2, 3){0}}
      \qbezier(86, 32)(80, 36.5)(83, 41)

      \qbezier(96, 34)(100, 36.5)(95.5, 43.5)
      \put(93.5, 31.5){\vector(-1, -1){0}}
      
      \put(82, -8){\vector(1, 2){6}}
      \put(72.5, -13){{\small $h_{t-1}^{l}$}}
      \put(98, -8){\vector(-1, 2){6}}
      \put(99, -13){{\small $h_{t}^{l-1}$}}

      \put(30, 87){\circle{16}}
      \put(28, 85){{\small $i$}}
      \put(6, 85){{\scriptsize $\begin{matrix}\text{Input}\\\text{gate}\end{matrix}$}}

      \put(21.5, 105){\vector(1, -2){5}}
      \put(12.5, 108){{\small $h_{t-1}^{l}$}}
      \put(39.5, 107){\vector(-1, -2){6}}
      \put(37, 108){{\small $h_{t}^{l-1}$}}

      \put(147, 87){\circle{16}}
      \put(144.5, 85){{\small $o$}}
      \put(156, 85){{\scriptsize $\begin{matrix}\text{Output}\\\text{gate}\end{matrix}$}}
        
      \put(138.5, 105){\vector(1, -2){5}}
      \put(129.5, 108){{\small $h_{t-1}^{l}$}}
      \put(156.5, 107){\vector(-1, -2){6}}
      \put(154, 108){{\small $h_{t}^{l-1}$}}

      \put(17, 50){\circle{16}}
      \put(15, 48){{\small $g$}}
      \put(1, 28){{\scriptsize $\begin{matrix}\text{Input}\\\text{modulation}\\\text{gate}\end{matrix}$}}

      \put(53.5, 50){\circle{6.5}}
      \put(50.75, 48.5){{\tiny $\times$}}

      \put(57, 50){\vector(1, 0){25}}
      \put(25, 50){\vector(1, 0){25}}
      \put(35, 80){\vector(2, -3){17.5}}

      \put(147, 50){\circle{6.5}}
      \put(144.25, 48.5){{\tiny $\times$}}
      \put(98, 50){\vector(1, 0){45.25}}
      \put(150.5, 50){\vector(1, 0){38}}

      \put(147, 79){\vector(0, -1){25.5}}
      \put(190, 47){${\mathbf h^l_t}$}

      \put(-20, 40){{\small $h_{t-1}^{l}$}}
      \put(-20, 56){{\small $h_{t}^{l-1}$}}
      \put(-10, 44){\vector(4, 1){19}}
      \put(-10, 58){\vector(4, -1){19}}

    \end{picture}
  \end{center}
  \caption{A graphical representation of LSTM memory cells used in this paper (there are minor differences in comparison to \citet{graves2013generating}).}
  \label{fig:lstm}
\end{figure}

The LSTM has complicated dynamics that allow it to
easily ``memorize'' information for an extended number of timesteps.  The
``long term'' memory is stored in a vector of \emph{memory cells}
$c^l_t \in \mathbb{R}^n$.  Although many LSTM architectures
that differ in their connectivity structure and activation functions,
all LSTM architectures have explicit memory cells for storing
information for long periods of time.  The LSTM can decide
to overwrite the memory cell, retrieve it, or keep it for the next time
step.  The LSTM architecture used in our experiments is given by the
following equations \cite{graves2013speech}:
\begin{align*}
&\text{LSTM} : h^{l-1}_t, h^l_{t-1}, c^l_{t - 1} \rightarrow h^l_t, c^l_t\\
&\begin{pmatrix}i\\f\\o\\g\end{pmatrix} =
  \begin{pmatrix}\mathrm{sigm}\\\mathrm{sigm}\\\mathrm{sigm}\\\tanh\end{pmatrix}
  T_{2n,4n}\begin{pmatrix}h^{l - 1}_t\\h^l_{t-1}\end{pmatrix}\\
&c^l_t = f \odot c^l_{t-1} + i \odot g\\
&h^l_t = o \odot \tanh(c^l_t)\\
\end{align*}
In these equations, $\mathrm{sigm}$ and $\tanh$ are applied
element-wise. Figure \ref{fig:lstm} illustrates the LSTM
equations.

\subsection{Regularization with Dropout} 
\label{sec:reg}

\begin{figure}
  \begin{center}
    \begin{picture}(150, 200)
      \multiput(0,0)(35, 0){6}{
        \put(-25, 45){\vector(1, 0){25}}
        \put(-25, 100){\vector(1, 0){25}}
      }
      \multiput(0,0)(35, 0){5}{
        \put(0, 0){
          \put(0, 85){\framebox(10, 30){}}
          \put(0, 30){\framebox(10, 30){}}
          \multiput(0,0)(0, 5){4}{
            \put(5, 7){\line(0, 0){2}}
            \put(5, 60){\line(0, 0){2}}
            \put(5, 115){\line(0, 0){2}}
          }
          \put(5, 30){\vector(0, 0){0.1}}
          \put(5, 85){\vector(0, 0){0.1}}
          \put(5, 138){\vector(0, 0){0.1}}
        }
      }
      \put(-2, 0){\makebox{$x_{t-2}$}}
      \put(33, 0){\makebox{$x_{t-1}$}}
      \put(71, 0){\makebox{$x_{t}$}}
      \put(103, 0){\makebox{$x_{t+1}$}}
      \put(138, 0){\makebox{$x_{t+2}$}}
      \put(-2, 142){\makebox{$y_{t-2}$}}
      \put(33, 142){\makebox{$y_{t-1}$}}
      \put(71, 142){\makebox{$y_{t}$}}
      \put(103, 142){\makebox{$y_{t+1}$}}
      \put(138, 142){\makebox{$y_{t+2}$}}
    \end{picture}
  \end{center}
  \caption{Regularized multilayer RNN. The dashed arrows indicate connections where dropout is applied, and
    the solid lines indicate connections where dropout is not applied.}
  \label{fig:reg}
\end{figure}
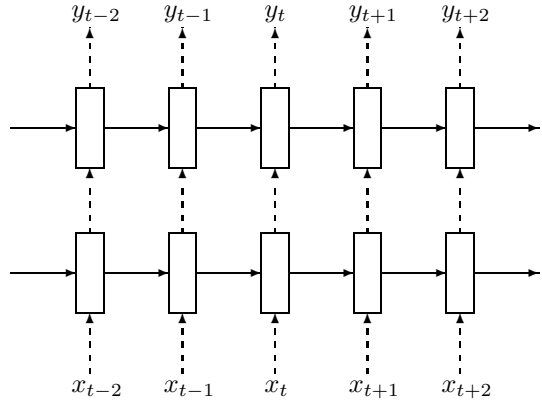

The main contribution of this paper is a recipe for applying 
dropout to LSTMs in a way that successfully reduces overfitting.
The main idea is to apply the dropout operator only to the non-recurrent connections
(Figure \ref{fig:reg}).  The following equation describes it more precisely,
where ${\bf D}$ is the dropout operator that sets a random subset of
its argument to zero:

\begin{align*}
&\begin{pmatrix}i\\f\\o\\g\end{pmatrix} =
  \begin{pmatrix}\mathrm{sigm}\\\mathrm{sigm}\\\mathrm{sigm}\\\tanh\end{pmatrix}
  T_{2n,4n}\begin{pmatrix}{\bf D}(h^{l - 1}_t)\\h^l_{t-1}\end{pmatrix}\\
&c^l_t = f \odot c^l_{t-1} + i \odot g\\
&h^l_t = o \odot \tanh(c^l_t)\\
\end{align*}

\begin{figure}
  \begin{center}
    \begin{picture}(150, 200)
      \multiput(0,0)(35, 0){6}{
        \put(-25, 45){\vector(1, 0){25}}
        \put(-25, 100){\vector(1, 0){25}}
      }
      \multiput(0,0)(35, 0){5}{
        \put(0, 0){
          \put(0, 85){\framebox(10, 30){}}
          \put(0, 30){\framebox(10, 30){}}
          \multiput(0,0)(0, 5){4}{
            \put(5, 7){\line(0, 0){2}}
            \put(5, 60){\line(0, 0){2}}
            \put(5, 115){\line(0, 0){2}}
          }
          \put(5, 30){\vector(0, 0){0.1}}
          \put(5, 85){\vector(0, 0){0.1}}
          \put(5, 138){\vector(0, 0){0.1}}
        }
      }
      \put(-2, 0){\makebox{$x_{t-2}$}}
      \put(33, 0){\makebox{$x_{t-1}$}}
      \put(71, 0){\makebox{$x_{t}$}}
      \put(103, 0){\makebox{$x_{t+1}$}}
      \put(138, 0){\makebox{$x_{t+2}$}}
      \put(-2, 142){\makebox{$y_{t-2}$}}
      \put(33, 142){\makebox{$y_{t-1}$}}
      \put(71, 142){\makebox{$y_{t}$}}
      \put(103, 142){\makebox{$y_{t+1}$}}
      \put(138, 142){\makebox{$y_{t+2}$}}

      {\linethickness{0.6mm}
        \put(5, 7){\line(0, 0){38}}
        \put(4, 45){\line(1, 0){105}}
        \put(110, 44){\line(0, 0){56}}
        \put(109, 100){\line(1, 0){35}}
        \put(145, 99){\line(0, 0){35}}
      }
    \end{picture}
  \end{center}
  \caption{The thick line shows a typical path of information flow in the LSTM. The
    information is affected by dropout $L + 1$ times, where $L$ is
    depth of network.}
  \label{fig:flow}
\end{figure}
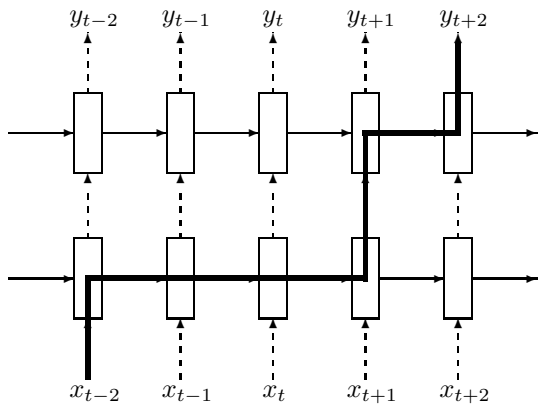

Our method works as follows. The dropout operator
corrupts the information carried by the units, forcing them to
perform their intermediate computations more robustly. At the same
time, we do not want to erase all the information from the units. It is especially
important that the units remember events that occurred many timesteps in the
past. Figure \ref{fig:flow} shows how information could flow from an
event that occurred at timestep ${t-2}$ to the prediction in timestep $t+2$
in our implementation of dropout. We can see that the information is
corrupted by the dropout operator exactly $L + 1$ times, and this
number is independent of the number of timesteps traversed by the
information.  Standard dropout perturbs the
recurrent connections, which makes it 
difficult for the LSTM to learn to store information for long periods of time.  By not
using dropout on the recurrent connections, the LSTM can benefit
from dropout regularization without sacrificing its valuable
memorization ability.

\section{Experiments}

We present results in three domains: language modeling (Section \ref{sec:lang}), 
speech recognition (Section \ref{sec:speech}), machine translation (Section \ref{sec:trans}),
and image caption generation (Section \ref{sec:caption}).

\subsection{Language modeling}
\label{sec:lang}

We conducted word-level prediction experiments on the Penn Tree Bank
(PTB) dataset \cite{marcus1993building}, which consists of $929$k
training words, $73$k validation words, and $82$k test words. It has
$10$k words in its vocabulary. We downloaded it from Tomas Mikolov's webpage\footnote{\url{http://www.fit.vutbr.cz/~imikolov/rnnlm/simple-examples.tgz}}.
We trained regularized LSTMs of two
sizes; these are denoted the medium LSTM and large LSTM.  Both LSTMs
have two layers and are unrolled for $35$ steps. We initialize the
hidden states to zero.  We then use the final hidden states of
the current minibatch as the initial hidden state of the subsequent minibatch
(successive minibatches sequentially traverse the training set).  
The size of each minibatch is 20.

\begin{table}[t]
  \small
  \centering
  \renewcommand{\arraystretch}{1.15}
  \begin{tabular}{lll}
    \hline
     Model & Validation set & Test set \\
    \hline
    \multicolumn{3}{c}{A single model} \\
    \hline
    \citet{pascanu2013construct} & & 107.5 \\
    \citet{chenglanguage} & & 100.0 \\
    non-regularized LSTM & 120.7 & 114.5 \\
    Medium regularized LSTM & 86.2 & 82.7 \\
    Large regularized LSTM & 82.2 & {\bf 78.4} \\
    \hline
    \multicolumn{3}{c}{Model averaging} \\
    \hline
    \citet{mikolov2012statistical} & & 83.5 \\
    \citet{chenglanguage} & & 80.6 \\
    2 non-regularized LSTMs & 100.4 & 96.1 \\
    5 non-regularized LSTMs & 87.9 & 84.1 \\
    10 non-regularized LSTMs & 83.5 & 80.0 \\
    2 medium regularized LSTMs & 80.6 & 77.0 \\
    5 medium regularized LSTMs & 76.7 & 73.3 \\
    10 medium regularized LSTMs & 75.2 & 72.0 \\
    2 large regularized LSTMs & 76.9 & 73.6 \\
    10 large regularized LSTMs & 72.8 & 69.5 \\
    38 large regularized LSTMs & 71.9 & {\bf 68.7} \\
    \hline
    \multicolumn{3}{c}{Model averaging with dynamic RNNs and n-gram models} \\
    \hline
    \citet{mikolov2012context} & & 72.9 \\
    \hline
  \end{tabular}
  \caption{Word-level perplexity on the Penn Tree Bank dataset.}
  \label{tab:ptb}
\end{table}

\begin{figure}
\line(1,0){400}

  {\footnotesize
  \textit{the meaning of life is} that only if an end would be of the whole supplier. widespread rules are regarded as the companies of refuses to deliver. in balance of the nation 's information and loan growth associated with the carrier thrifts are in the process of slowing the seed and commercial paper.}
\line(1,0){400}

  {\footnotesize
\textit{the meaning of life is} nearly in the first several months before the government was addressing such a move as president and chief executive of the nation past from a national commitment to curb grounds. meanwhile the government invests overcapacity that criticism and in the outer reversal of small-town america.}

\line(1,0){400}
  \caption{Some interesting samples drawn from a large regularized model conditioned on ``The meaning of life is''. We have removed ``unk'', ``N'', ``\$'' from the set of permissible words.}
  \label{fig:meaning}
\end{figure}

The medium LSTM has $650$ units per layer
and its parameters are initialized uniformly in $[-0.05,
  0.05]$. As described earlier, we apply $50\%$ dropout on the non-recurrent connections. We
train the LSTM for $39$ epochs with a learning rate of $1$, and after
$6$ epochs we decrease it by a factor of $1.2$ after each epoch. We
clip the norm of the gradients (normalized by minibatch size) at $5$. 
Training this network takes about half a day on an NVIDIA K20 GPU. 

The large LSTM has $1500$ units per layer and its parameters are
initialized uniformly in $[-0.04, 0.04]$. We apply $65\%$ dropout on
the non-recurrent connections. We train the model for $55$ epochs with
a learning rate of $1$;  after $14$ epochs we start to reduce the learning rate by a
factor of $1.15$ after each epoch. We clip the norm of
the gradients (normalized by minibatch size) at $10$ \cite{mikolov2010recurrent}. Training this
network takes an entire day on an NVIDIA K20 GPU.

For comparison, we trained a non-regularized network. 
We optimized its parameters to get the best validation performance.
The lack of regularization
effectively constrains size of the network, forcing us to use small network because larger networks overfit. 
Our best performing non-regularized LSTM has two hidden layers with $200$ units per layer, and its weights are initialized
uniformly in $[-0.1, 0.1]$.
We train it for $4$ epochs 
with a learning rate of $1$ and then we decrease the learning rate by a factor of $2$ after each epoch, for a total of $13$ training epochs.
The size of each minibatch is $20$, and we unroll the network for $20$ steps. Training this network takes 2-3 hours
on an NVIDIA K20 GPU.

Table \ref{tab:ptb} compares previous results with our LSTMs, and
Figure \ref{fig:meaning} shows samples drawn from a single large 
regularized LSTM.

\subsection{Speech recognition}
\label{sec:speech}

Deep Neural Networks have been used for acoustic modeling for over half a century (see
\citet{BourlardASR} for a good review). Acoustic modeling is a key
component in mapping acoustic signals to sequences of words, as it
models $p(s_t|X)$ where $s_t$ is the phonetic state at time $t$ and $X$
is the acoustic observation. Recent work has shown that LSTMs can 
achieve excellent performance on acoustic modeling \cite{sak2014speech}, yet
relatively small LSTMs (in terms of the number of their parameters) can
easily overfit the training set. A useful metric for measuring the performance of acoustic models is frame
accuracy, which is measured at each $s_t$ for all timesteps
$t$. Generally, this metric correlates with the actual metric of
interest, the Word Error Rate (WER). Since computing the WER
involves using a language model and tuning the decoding parameters for
every change in the acoustic model, we decided to focus on frame
accuracy in these experiments. Table~\ref{tab:speech} shows
that dropout improves the frame accuracy of the LSTM. Not
surprisingly, the training frame accuracy drops due to the noise added
during training, but as is often the case with dropout, this yields
models that generalize better to unseen data. Note that the test
set is easier than the training set, as its accuracy is higher.  We
report the performance of an LSTM on an internal Google Icelandic
Speech dataset, which is relatively small (93k utterances), so
overfitting is a great concern.

\begin{table}[t]
  \small
  \centering
  \renewcommand{\arraystretch}{1.15}
  \begin{tabular}{lll}
    \hline
     Model & Training set & Validation set \\
    \hline
    Non-regularized LSTM & 71.6 & 68.9 \\
    Regularized LSTM & 69.4 & {\bf 70.5} \\
    \hline
  \end{tabular}
  \caption{Frame-level accuracy on the Icelandic Speech Dataset. The training set has 93k utterances.}
  \label{tab:speech}
\end{table}

\subsection{Machine translation}
\label{sec:trans}

We formulate a machine translation problem as a language modelling task, where
an LSTM is trained to assign high probability to a correct
translation of a source sentence.  Thus, the LSTM is trained on
concatenations of source sentences and their translations 
\cite{sutskever2014sequence} (see also \cite{cho2014learning}). We compute a translation by 
approximating the most probable sequence of words
using a simple beam search with a beam of size 12.  We ran an
LSTM on the WMT'14 English to French dataset, on the
``selected'' subset from \citet{wmt_joint} which has 340M French words
and 304M English words.  Our LSTM has 4 hidden layers, and both its
layers and word embeddings have 1000 units.  Its 
English vocabulary has 160,000 words and its French vocabulary has
80,000 words.  The optimal dropout probability was 0.2.
Table \ref{tab:mt} shows the performance of an LSTM trained with and without dropout.
While our LSTM does not beat the phrase-based LIUM SMT system
\cite{lium}, our results show that dropout improves the
translation performance of the LSTM.

\begin{table}[t]
  \small
  \centering
  \renewcommand{\arraystretch}{1.15}
  \begin{tabular}{lll}
    \hline
     Model & Test perplexity & Test BLEU score \\
    \hline
    Non-regularized LSTM & 5.8 & 25.9 \\
    Regularized LSTM & 5.0 &  29.03 \\
    \hline
    LIUM system  &  &  33.30 \\
    \hline
  \end{tabular}
  \caption{Results on the English to French translation task. }
  \label{tab:mt}
\end{table}

\subsection{Image Caption Generation}
\label{sec:caption}

We applied the dropout variant 
to the image caption generation model of \cite{vinyals2014show}.  The 
image caption generation is similar to the sequence-to-sequence model
of \cite{sutskever2014sequence}, but where the input image is mapped
onto a vector with a highly-accurate pre-trained convolutional
neural network \citep{szegedy2014going}, which is converted
into a caption with a single-layer LSTM (see \citet{vinyals2014show}
for the details on the architecture).  We test our dropout scheme on  
LSTM as the convolutional neural network is not trained on the image caption
dataset because it is not large (MSCOCO \citep{lin2014microsoft}). 

Our results are summarized in the following Table \ref{tab:vis}.  In brief,
dropout helps relative to not using dropout, but using an ensemble
eliminates the gains attained by dropout.  Thus, in this setting,
the main effect of dropout is to produce a single model that is as
good as an ensemble, which is a reasonable improvement given
the simplicity of the technique.  
\begin{table}[t]
  \small
  \centering
  \renewcommand{\arraystretch}{1.15}
  \begin{tabular}{lll}
    \hline
     Model & Test perplexity & Test BLEU score \\
    \hline
    Non-regularized model & 8.47 & 23.5 \\
    Regularized model & 7.99 &  24.3 \\
    \hline
    10 non-regularized models  & 7.5  &  24.4 \\
    \hline
  \end{tabular}
  \caption{Results on the image caption generation task. }
  \label{tab:vis}
\end{table}

\section{Conclusion}

We presented a simple way of applying dropout to LSTMs that results in
large performance increases on several problems in different domains.
Our work makes dropout useful for RNNs, and our results suggest that
our implementation of dropout could improve performance on a wide
variety of applications.

\section{Acknowledgments}

We wish to acknowledge Tomas Mikolov for useful comments on the first version of the paper.   

\small
\bibliography{bibliography}
\bibliographystyle{icml2014}

\end{document}